\documentclass[11pt,letterpaper]{article}
\usepackage{fullpage} 
\usepackage{natbib}
\usepackage{graphicx}
\usepackage[hyphens]{url}
\usepackage{color}
\begin{document}
\newcommand{\changed}[1]{\textcolor{blue}{#1}}

\title{{Towards A Rigorous Science of Interpretable Machine Learning}}
\author{Finale Doshi-Velez$^{*}$ and Been Kim\footnote{Authors contributed equally.}} 
\date{} 
\maketitle

From autonomous cars and adaptive email-filters to predictive policing
systems, machine learning (ML) systems are increasingly ubiquitous;
they outperform humans on specific tasks
\citep{mnih2013playing,silver2016mastering,cmupoker} and often guide
processes of human understanding and decisions
\citep{carton2016identifying, doshi2014comorbidity}.  The deployment
of ML systems in complex applications has led to a surge of interest
in systems optimized not only for expected task performance but also
other important criteria such as safety \citep{otte2013safe,
  amodei2016concrete, VarshneyA16}, nondiscrimination
\citep{bostrom2014ethics, ruggieri2010data, hardt2016equality},
avoiding technical debt \citep{sculley2015hidden}, or providing the
right to explanation \citep{goodman2016european}.  For ML systems to
be used safely, satisfying these auxiliary criteria is critical.
However, unlike measures of performance such as accuracy, these
criteria often cannot be completely quantified. For example, we might
not be able to enumerate all unit tests required for the safe
operation of a semi-autonomous car or all confounds that might cause a
credit scoring system to be discriminatory.  In such cases, a popular
fallback is the criterion of \emph{interpretability}: if the system
can \emph{explain} its reasoning, we then can verify whether that
reasoning is sound with respect to these auxiliary criteria.

Unfortunately, there is little consensus on what interpretability in
machine learning \emph{is} and how to \emph{evaluate} it for
benchmarking.  Current interpretability evaluation typically falls
into two categories. The first evaluates interpretability in the
context of an application: if the system is useful in either a
practical application or a simplified version of it, then it must be
somehow interpretable (e.g. \citet{ribeiro2016should,
  lei2016rationalizing, kim2015ibcm, doshi2014graph, kim2015mind}).
The second evaluates interpretability via a quantifiable proxy: a
researcher might first claim that some model class---e.g. sparse
linear models, rule lists, gradient boosted trees---are interpretable
and then present algorithms to optimize within that class
(e.g. \citet{buciluǎ2006model, bayesian2017wang, wang2015falling,
  lou2012intelligible}).

To large extent, both evaluation approaches rely on some notion of
``you'll know it when you see it.'' Should we be concerned about a
lack of rigor?  Yes and no: the notions of interpretability above
appear reasonable because they \emph{are} reasonable: they meet the
first test of having face-validity on the correct test set of
subjects: human beings.  However, this basic notion leaves many kinds
of questions unanswerable: Are all models in all
defined-to-be-interpretable model classes equally interpretable?
Quantifiable proxies such as sparsity may seem to allow for
comparison, but how does one think about comparing a model sparse in
features to a model sparse in prototypes?  Moreover, do all
applications have the same interpretability needs?  If we are to move
this field forward---to compare methods and understand when methods
may generalize---we need to formalize these notions and make them
evidence-based.

\begin{figure}
\centering
\includegraphics[scale=.3]{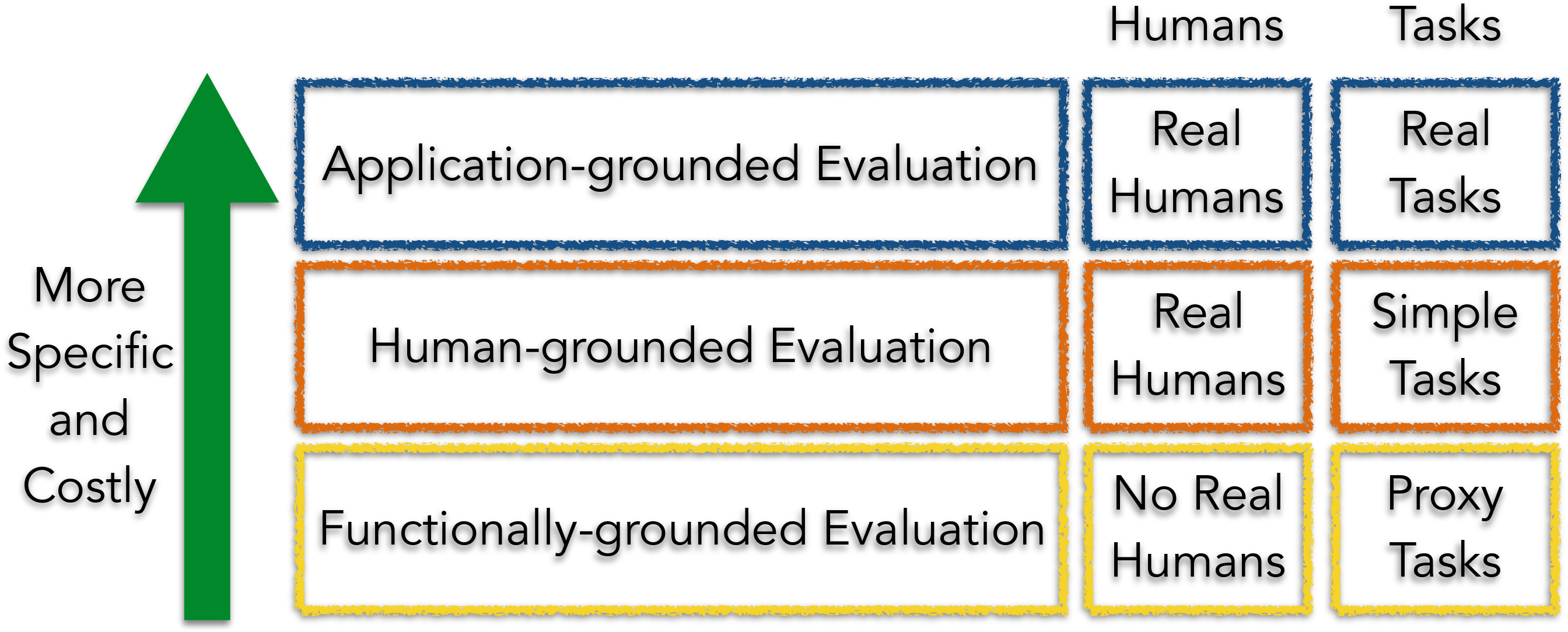}
\caption{Taxonomy of evaluation approaches for interpretability}
\end{figure}

The objective of this review is to chart a path toward the definition
and rigorous evaluation of interpretability. The need is urgent:
recent European Union regulation will \emph{require} algorithms
that
make decisions based on user-level predictors, which
"significantly affect" users to
provide explanation (``right to explanation'') by 2018~\citep{eu_reg}.
In addition, the volume of research on interpretability is rapidly
growing.\footnote{Google Scholar finds more than 20,000 publications
  related to interpretability in ML in the last five years.}  In
section~\ref{sec:what}, we discuss what interpretability is and
contrast with other criteria such as reliability and fairness. In
section~\ref{sec:why}, we consider scenarios in which interpretability
is needed and why.  In section~\ref{sec:how}, we propose a taxonomy
for the evaluation of interpretability---application-grounded,
human-grounded and functionally-grounded.  We conclude with important
open questions in section~\ref{sec:links} and specific suggestions for
researchers doing work in interpretability in section~\ref{sec:conc}.

\section{What is Interpretability?}
\label{sec:what} 

\paragraph{Definition} 
Interpret means \emph{to explain or to present in understandable
  terms}.\footnote{Merriam-Webster dictionary, accessed 2017-02-07} In
the context of ML systems, we define interpretability as the
\emph{ability to explain or to present in understandable terms to a
  human}.  A formal definition of explanation remains elusive; in the
field of psychology, \citet{lombrozo2006structure} states
``explanations are... the currency in which we exchanged beliefs'' and
notes that questions such as what constitutes an explanation, what
makes some explanations better than others, how explanations are
generated and when explanations are sought are just beginning to be
addressed.  Researchers have classified explanations from being
``deductive-nomological" in nature \citep{hempel1948studies} (i.e. as
logical proofs) to providing some sense of mechanism
\citep{bechtel2005explanation, chater2006speculations,
  glennan2002rethinking}.  \citet{keil2006explanation} considered a
broader definition: implicit explanatory understanding.
In this work, we propose data-driven ways to derive operational
definitions and evaluations of explanations, and thus,
interpretability.

\paragraph{Interpretability is used to confirm other important desiderata of ML systems}
There exist many auxiliary criteria that one may wish to optimize.
Notions of \emph{fairness} or \emph{unbiasedness} imply that protected
groups (explicit or implicit) are not somehow discriminated against.
\emph{Privacy} means the method protects sensitive information in the
data.  Properties such as \emph{reliability} and \emph{robustness}
ascertain whether algorithms reach certain levels of performance in
the face of parameter or input variation.  \emph{Causality} implies
that the predicted change in output due to a perturbation will occur
in the real system.  \emph{Usable} methods provide information that
assist users to accomplish a task---e.g. a knob to tweak image
lighting---while \emph{trusted} systems have the confidence of human
users---e.g. aircraft collision avoidance systems.  Some areas, such
as the fairness \citep{hardt2016equality} and privacy
\citep{toubiana2010adnostic, Dwork2012, Hardt2010} the research
communities have formalized their criteria, and these formalizations
have allowed for a blossoming of rigorous research in these fields
(without the need for interpretability).  However, in many cases,
formal definitions remain elusive.  Following the psychology
literature, where \citet{keil2004lies} notes ``explanations may
highlight an incompleteness,''
we argue that interpretability can assist in qualitatively
ascertaining whether other desiderata---such as fairness, privacy,
reliability, robustness, causality, usability and trust---are met.
For example, one can provide a feasible explanation that fails to
correspond to a causal structure, exposing a potential concern. 

\section{Why interpretability? Incompleteness} 
\label{sec:why} 
Not all ML systems require interpretability.  Ad servers, postal code
sorting, air craft collision avoidance systems---all compute their
output without human intervention.  Explanation is not necessary
either because (1) there are no significant consequences for
unacceptable results or (2) the problem is sufficiently well-studied
and validated in real applications that we trust the system's
decision, even if the system is not perfect.

So when is explanation necessary and appropriate?  We argue that the
need for interpretability stems from an \emph{incompleteness} in the
problem formalization, creating a fundamental barrier to optimization
and evaluation.  Note that incompleteness is distinct from
uncertainty: the fused estimate of a missile location may be
uncertain, but such uncertainty can be rigorously quantified and
formally reasoned about.  In machine learning terms, we distinguish
between cases where unknowns result in quantified
variance---e.g. trying to learn from small data set or with limited
sensors---and incompleteness that produces some kind of unquantified
bias---e.g. the effect of including domain knowledge in a model
selection process.  Below are some illustrative scenarios:
\begin{itemize}
\item Scientific Understanding: The human's goal is to gain knowledge.
  We do not have a complete way of stating what knowledge is; thus the
  best we can do is ask for explanations we can convert into
  knowledge.
\item Safety: For complex tasks, the end-to-end system is almost never
  completely testable; one cannot create a complete list of scenarios
  in which the system may fail. Enumerating all possible outputs given
  all possible inputs be computationally or logistically infeasible,
  and we may be unable to flag all undesirable outputs.
\item Ethics: The human may want to guard against certain kinds of
  discrimination, and their notion of fairness may be too abstract to
  be completely encoded into the system (e.g., one might desire a
  `fair' classifier for loan approval).  Even if we can encode
  protections for specific protected classes into the system, there
  might be biases that we did not consider a priori (e.g., one may not
  build gender-biased word embeddings on purpose, but it was a pattern
  in data that became apparent only after the fact).
\item Mismatched objectives: The agent's algorithm may be optimizing
  an incomplete objective---that is, a proxy function for the ultimate
  goal.  For example, a clinical system may be optimized for
  cholesterol control, without considering the likelihood of
  adherence; an automotive engineer may be interested in engine data
  not to make predictions about engine failures but to more broadly
  build a better car.
\item Multi-objective trade-offs: Two well-defined desiderata in ML
  systems may compete with each other, such as privacy and prediction
  quality \citep{hardt2016equality} or privacy and non-discrimination
  \citep{strahilevitz2008privacy}.  Even if each objectives are
  fully-specified, the exact dynamics of the trade-off may not be
  fully known, and the decision may have to be case-by-case.
\end{itemize} 
In the presence of an incompleteness, explanations are one of ways to
ensure that effects of gaps in problem formalization are visible to us.

\section{How? A Taxonomy of Interpretability Evaluation} 
\label{sec:how} 
Even in standard ML settings, there exists a taxonomy of evaluation
that is considered appropriate.  In particular, the evaluation should
match the claimed contribution.  Evaluation of applied work should
demonstrate success in the application: a game-playing agent might
best a human player, a classifier may correctly identify star types
relevant to astronomers.  In contrast, core methods work should
demonstrate generalizability via careful evaluation on a variety of
synthetic and standard benchmarks.

In this section we lay out an analogous taxonomy of evaluation
approaches for interpretability: application-grounded, human-grounded,
and functionally-grounded. These range from task-relevant to general,
also acknowledge that while human evaluation is essential to assessing
interpretability, human-subject evaluation is not an easy task.  A
human experiment needs to be well-designed to minimize confounding
factors, consumed time, and other resources.  We discuss the
trade-offs between each type of evaluation and when each would be
appropriate.

\subsection{Application-grounded Evaluation: Real humans, real tasks} 
Application-grounded evaluation involves conducting human experiments
within a real application.  If the researcher has a concrete
application in mind---such as working with doctors on diagnosing
patients with a particular disease---the best way to show that the
model works is to evaluate it with respect to the task: doctors
performing diagnoses.  This reasoning aligns with the methods of
evaluation common in the human-computer interaction and visualization
communities, where there exists a strong ethos around making sure that
the system delivers on its intended task
\citep{antunes2012structuring, lazar2010research}.  For example, a
visualization for correcting segmentations from microscopy data would
be evaluated via user studies on segmentation on the target image task
\citep{suissa2016automatic}; a homework-hint system is evaluated on
whether the student achieves better post-test performance
\citep{williams2016axis}.

Specifically, we evaluate the quality of an explanation in the context
of its end-task, such as whether it results in better identification
of errors, new facts, or less discrimination.  Examples of experiments
include:
\begin{itemize}
\item Domain expert experiment with the exact application task.
\item Domain expert experiment with a simpler or partial task to
  shorten experiment time and increase the pool of potentially-willing
  subjects.
\end{itemize}
In both cases, an important baseline is how well \emph{human-produced}
explanations assist in other humans trying to complete the task.  To
make high impact in real world applications, it is essential that we
as a community respect the time and effort involved to do such
evaluations, and also demand high standards of experimental design
when such evaluations are performed.  As HCI community
recognizes~\citep{antunes2012structuring}, this is not an easy
evaluation metric.  Nonetheless, it directly tests the objective that
the system is built for, and thus performance with respect to that
objective gives strong evidence of success.

\subsection{Human-grounded Metrics: Real humans, simplified tasks}
Human-grounded evaluation is about conducting simpler human-subject
experiments that maintain the essence of the target application.  Such
an evaluation is appealing when experiments with the target community
is challenging. These evaluations can be completed with lay humans,
allowing for both a bigger subject pool and less expenses, since we do
not have to compensate highly trained domain experts.  Human-grounded
evaluation is most appropriate when one wishes to test more general
notions of the quality of an explanation.  For example, to study what
kinds of explanations are best understood under severe time
constraints, one might create abstract tasks in which other
factors---such as the overall task complexity---can be
controlled~\citep{kim2013inferring, lakkaraju2016interpretable}

The key question, of course, is how we can evaluate the quality of an
explanation without a specific end-goal (such as identifying errors in
a safety-oriented task or identifying relevant patterns in a
science-oriented task). Ideally, our evaluation approach will depend
only on the quality of the explanation, regardless of whether the
explanation is the model itself or a post-hoc interpretation of a
black-box model, and regardless of the correctness of the associated
prediction.  Examples of potential experiments include:
\begin{itemize}
\item Binary forced choice: humans are presented with pairs of
  explanations, and must choose the one that they find of higher
  quality (basic face-validity test made quantitative).
 \item Forward simulation/prediction: humans are presented with an
   explanation and an input, and must correctly simulate the model's
   output (regardless of the true output).  
\item Counterfactual simulation: humans are presented with an
  explanation, an input, and an output, and are asked what must be
  changed to change the method's prediction to a desired output (and
  related variants).  
\end{itemize}

Here is a concrete example. The common intrusion-detection
test~\citep{chang2009reading} in topic models is a form of the forward
simulation/prediction task: we ask the human to find the difference
between the model's true output and some corrupted output as a way to
determine whether the human has correctly understood what the model's
true output is.

\subsection{Functionally-grounded Evaluation: No humans, proxy tasks}
Functionally-grounded evaluation requires no human experiments;
instead, it uses some formal definition of interpretability as a proxy
for explanation quality.  Such experiments are appealing because even
general human-subject experiments require time and costs both to
perform and to get necessary approvals (e.g., IRBs), which may be
beyond the resources of a machine learning researcher.
Functionally-grounded evaluations are most appropriate once we have a
class of models or regularizers that have already been validated,
e.g. via human-grounded experiments.  They may also be appropriate
when a method is not yet mature or when human subject experiments are
unethical.

The challenge, of course, is to determine what proxies to use.  For
example, decision trees have been considered interpretable in many
situations~\citep{freitas2014comprehensible}.  In
section~\ref{sec:links}, we describe open problems in determining what
proxies are reasonable.  Once a proxy has been formalized, the
challenge is squarely an optimization problem, as the model class or
regularizer is likely to be discrete, non-convex and often
non-differentiable.  Examples of experiments include
\begin{itemize}
\item Show the improvement of prediction performance of a model that
  is already proven to be interpretable (assumes that someone has run
  human experiments to show that the model class is interpretable).
\item Show that one's method performs better with respect to certain
  regularizers---for example, is more sparse---compared to other
  baselines (assumes someone has run human experiments to show that
  the regularizer is appropriate).
\end{itemize}

\section{Open Problems in the Science of Interpretability, Theory and Practice}
\label{sec:links} 
It is essential that the three types of evaluation in the previous
section inform each other: the factors that capture the essential
needs of real world tasks should inform what kinds of simplified tasks
we perform, and the performance of our methods with respect to
functional proxies should reflect their performance in real-world
settings.  In this section, we describe some important open problems
for creating these links between the three types of evaluations:
\begin{enumerate}
\item What proxies are best for what real-world applications?
  (functionally to application-grounded)
\item What are the important factors to consider when designing 
simpler tasks that maintain the essence of the real end-task? (human to application-grounded)
\item What are the important factors to consider when characterizing
  proxies for explanation quality? (human to functionally-grounded)
\end{enumerate}  
Below, we describe a path to answering each of these questions.

\subsection{Data-driven approach to discover factors of interpretability}
\begin{figure*}
\centering
\includegraphics[scale=0.2]{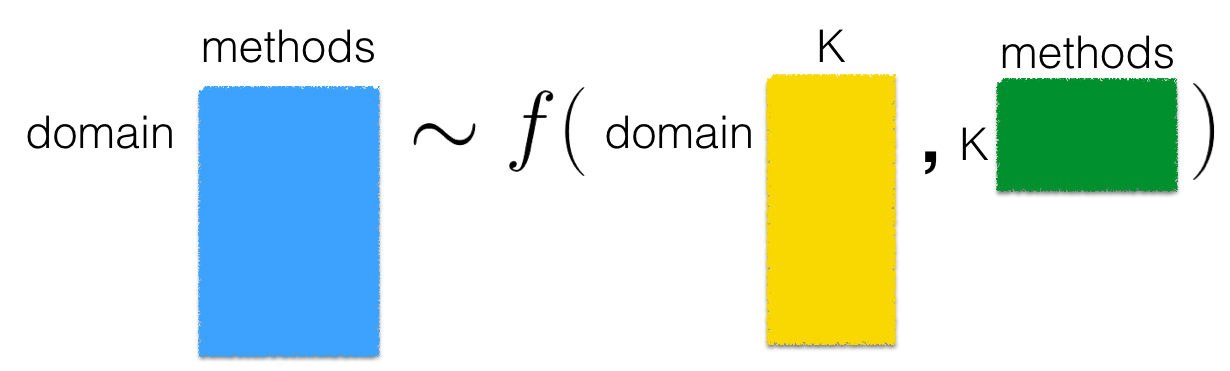}
\caption{An example of data-driven approach to discover factors in
interpretability\label{fig:coallb}}
\end{figure*}

Imagine a matrix where rows are specific real-world tasks, columns are
specific methods, and the entries are the performance of the method on
the end-task. For example, one could represent how well a decision
tree of depth less than 4 worked in assisting doctors in identifying
pneumonia patients under age 30 in US. Once constructed, methods in
machine learning could be used to identify latent dimensions that
represent factors that are important to interpretability.  This
approach is similar to efforts to characterize
classification \citep{ho2002complexity} and clustering problems
\citep{garg2016meta}.  For example, one might perform matrix
factorization to embed both tasks and methods respectively in
low-dimensional spaces (which we can then seek to interpret), as shown
in Figure~\ref{fig:coallb}.  These embeddings could help predict what
methods would be most promising for a new problem, similarly to
collaborative filtering.

The challenge, of course, is in creating this matrix.  For example,
one could imagine creating a repository of clinical cases in which the
ML system has access to the patient's record but not certain current
features that are only accessible to the clinician, or a repository of
discrimination-in-loan cases where the ML system must provide outputs
that assist a lawyer in their decision.  Ideally these would be linked
to domain experts who have agreed to be employed to evaluate methods
when applied to their domain of expertise.  \emph{Just as there are
  now large open repositories for problems in classification,
  regression, and reinforcement learning \citep{blake1998uci,
    brockman2016openai, vanschoren2014openml}, we advocate for the
  creation of repositories that contain problems corresponding to
  real-world tasks in which human-input is required.}  Creating such
repositories will be more challenging than creating collections of
standard machine learning datasets because they must include a system
for human assessment, but with the availablity of crowdsourcing tools
these technical challenges can be surmounted.  

In practice, constructing such a matrix will be expensive since each
cell must be evaluated in the context of a real application, and
interpreting the latent dimensions will be an iterative effort of
hypothesizing why certain tasks or methods share dimensions and then
checking whether our hypotheses are true.  In the next two open
problems, we lay out some hypotheses about what latent dimensions may
correspond to; these hypotheses can be tested via much less expensive
human-grounded evaluations on simulated tasks.

\subsection{Hypothesis: task-related latent dimensions of interpretability}
\label{sec:task_type}
Disparate-seeming applications may share common categories: an
application involving preventing medical error at the bedside and an
application involving support for identifying inappropriate language
on social media might be similar in that they involve making a
decision about a specific case---a patient, a post---in a relatively
short period of time.  However, when it comes to time constraints, the
needs in those scenarios might be different from an application
involving the understanding of the main characteristics of a large
omics data set, where the goal---science---is much more abstract and
the scientist may have hours or days to inspect the model outputs.

Below, we list a (non-exhaustive!) set of hypotheses about what might
make tasks similar in their explanation needs: 
\begin{itemize}
\item \emph{Global vs. Local.} 
  Global interpretability implies knowing what patterns are present in
  general (such as key features governing galaxy formation), while
  local interpretability implies knowing the reasons for a specific
  decision (such as why a particular loan application was rejected).
  The former may be important for when scientific understanding or
  bias detection is the goal; the latter when one needs a
  justification for a specific decision.
\item \emph{Area, Severity of Incompleteness.}  What part of the
  problem formulation is incomplete, and how incomplete is it?  We
  hypothesize that the types of explanations needed may vary depending
  on whether the source of concern is due to incompletely specified
  inputs, constraints, domains, internal model structure, costs, or
  even in the need to understand the training algorithm.  The severity
  of the incompleteness may also affect explanation needs.  For
  example, one can imagine a spectrum of questions about the safety of
  self-driving cars.  On one end, one may have general curiosity about
  how autonomous cars make decisions.  At the other, one may wish to
  check a specific list of scenarios (e.g., sets of sensor inputs that
  causes the car to drive off of the road by 10cm). In between, one
  might want to check a general property---safe urban
  driving---without an exhaustive list of scenarios and safety
  criteria.
\item \emph{Time Constraints.}  How long can the user afford to spend
  to understand the explanation?  A decision that needs to be made at
  the bedside or during the operation of a plant must be understood
  quickly,
  while in scientific or anti-discrimination applications, the
  end-user may be willing to spend hours trying to fully understand an
  explanation.
\item \emph{Nature of User Expertise.}  How experienced is the user in
  the task?  The user's experience will affect what kind of
  \emph{cognitive chunks} they have, that is, how they organize
  individual elements of information into collections
  \citep{neath2003human}.  For example, a clinician may have a notion
  that autism and ADHD are both developmental diseases.  The nature of
  the user's expertise will also influence what level of
  sophistication they expect in their explanations. For example,
  domain experts may expect or prefer a somewhat larger and
  sophisticated model---which confirms facts they know---over a
  smaller, more opaque one.  These preferences may be quite different
  from hospital ethicist who may be more narrowly concerned about
  whether decisions are being made in an ethical manner.  More
  broadly, decison-makers, scientists, compliance and safety
  engineers, data scientists, and machine learning researchers all
  come with different background knowledge and communication styles.
\end{itemize}
Each of these factors can be isolated in human-grounded experiments in
simulated tasks to determine which methods work best when they are
present.

\subsection{Hypothesis: method-related latent dimensions of interpretability}
\label{sec:method_type}
Just as disparate applications may share common categories, disparate
methods may share common qualities that correlate to their utility as
explanation.  As before, we provide a (non-exhaustive!) set of factors
that may correspond to different explanation needs: Here, we define
\emph{cognitive chunks} to be the basic units of explanation.

\begin{itemize}

\item \emph{Form of cognitive chunks.}  What are the basic units of
  the explanation?  Are they raw features?  Derived features that have
  some semantic meaning to the expert (e.g. ``neurological disorder''
  for a collection of diseases or ``chair'' for a collection of
  pixels)?  Prototypes?  
\item \emph{Number of cognitive chunks.}  How many cognitive chunks
  does the explanation contain?  How does the quantity interact with
  the type: for example, a prototype can contain a lot more
  information than a feature; can we handle them in similar
  quantities?
\item \emph{Level of compositionality.}  Are the cognitive chunks organized
  in a structured way?  Rules, hierarchies, and other abstractions can
  limit what a human needs to process at one time.  For example, part
  of an explanation may involve \emph{defining} a new unit (a chunk) that is a
  function of raw units, and then providing an explanation in terms of
  that new unit.
\item \emph{Monotonicity and other interactions between cognitive
  chunks.}  Does it matter if the cognitive chunks are combined in
  linear or nonlinear ways?  In monotone ways
  \citep{gupta2016monotonic}?  Are some functions more natural to
  humans than others \citep{wilson2015human, schulz2016compositional}?
\item \emph{Uncertainty and stochasticity.}  How well do people
  understand uncertainty measures? To what extent is stochasticity
  understood by humans?
\end{itemize} 

\section{Conclusion: Recommendations for Researchers} 
\label{sec:conc} 
In this work, we have laid the groundwork for a process to rigorously
define and evaluate interpretability.  There are many open questions
in creating the formal links between applications, the science of
human understanding, and more traditional machine learning
regularizers.  In the mean time, we encourage the community to
consider some general principles.

\emph{The claim of the research should match the type of the
  evaluation.}  Just as one would be critical of a
reliability-oriented paper that only cites accuracy statistics, the
choice of evaluation should match the specificity of the claim being
made.  A contribution that is focused on a particular application
should be expected to be evaluated in the context of that application
(application-grounded evaluation), or on a human experiment with a
closely-related task (human-grounded evaluation).  A contribution that
is focused on better optimizing a model class for some definition of
interpretability should be expected to be evaluated with
functionally-grounded metrics.  As a community, we must be careful in
the work on interpretability, both recognizing the need for and the
costs of human-subject experiments.

\emph{We should categorize our applications and methods with a common
  taxonomy.}  In section~\ref{sec:links}, we hypothesized factors that
may be the latent dimensions of interpretability.  Creating a shared
language around such factors is essential not only to evaluation, but
also for the citation and comparison of related work.  For example,
work on creating a safe healthcare agent might be framed as focused on
the need for explanation due to unknown inputs at the local scale,
evaluated at the level of an application.  In contrast, work on
learning sparse linear models might also be framed as focused on the
need for explanation due to unknown inputs, but this time evaluated at
global scale.  As we share each of our work with the community, we can
do each other a service by describing factors such as
\begin{enumerate}
\item How is the problem formulation incomplete?
  (Section~\ref{sec:why})
\item At what level is the evaluation being performed? (application,
  general user study, proxy; Section~\ref{sec:how})
\item What are task-related relevant factors? (e.g. global vs. local,
  severity of incompleteness, level of user expertise, time
  constraints; Section~\ref{sec:task_type})
\item What are method-related relevant factors being explored?
  (e.g. form of cognitive chunks, number of cognitive chunks,
  compositionality, monotonicity, uncertainty;
  Section~\ref{sec:method_type})
\end{enumerate}
and of course, adding and refining these factors as our taxonomies
evolve.  These considerations should move us away from vague claims
about the interpretability of a particular model and toward
classifying applications by a common set of terms.

\paragraph{Acknowledgments}  
This piece would not have been possible without the dozens of deep
conversations about interpretability with machine learning researchers
and domain experts. Our friends and colleagues, we appreciate your
support. We want to particularity thank Ian Goodfellow, Kush Varshney, Hanna Wallach,
Solon Barocas, Stefan Rüping and Jesse Johnson for their feedback.

\bibliographystyle{plainnat} 
\bibliography{main_from_Been}

\newcommand{\noopsort}[1]{} \newcommand{\printfirst}[2]{#1}
  \newcommand{\singleletter}[1]{#1} \newcommand{\switchargs}[2]{#2#1}
\begin{thebibliography}{51}
\providecommand{\natexlab}[1]{#1}
\providecommand{\url}[1]{\texttt{#1}}
\expandafter\ifx\csname urlstyle\endcsname\relax
  \providecommand{\doi}[1]{doi: #1}\else
  \providecommand{\doi}{doi: \begingroup \urlstyle{rm}\Url}\fi

\bibitem[Amodei et~al.(2016)Amodei, Olah, Steinhardt, Christiano, Schulman, and
  Man{\'e}]{amodei2016concrete}
Dario Amodei, Chris Olah, Jacob Steinhardt, Paul Christiano, John Schulman, and
  Dan Man{\'e}.
\newblock Concrete problems in {AI} safety.
\newblock \emph{arXiv preprint arXiv:1606.06565}, 2016.

\bibitem[Antunes et~al.(2012)Antunes, Herskovic, Ochoa, and
  Pino]{antunes2012structuring}
Pedro Antunes, Valeria Herskovic, Sergio~F Ochoa, and Jose~A Pino.
\newblock Structuring dimensions for collaborative systems evaluation.
\newblock \emph{ACM Computing Surveys}, 2012.

\bibitem[Bechtel and Abrahamsen(2005)]{bechtel2005explanation}
William Bechtel and Adele Abrahamsen.
\newblock Explanation: A mechanist alternative.
\newblock \emph{Studies in History and Philosophy of Science Part C: Studies in
  History and Philosophy of Biological and Biomedical Sciences}, 2005.

\bibitem[Blake and Merz(1998)]{blake1998uci}
Catherine Blake and Christopher~J Merz.
\newblock $\{$UCI$\}$ repository of machine learning databases.
\newblock 1998.

\bibitem[Bostrom and Yudkowsky(2014)]{bostrom2014ethics}
Nick Bostrom and Eliezer Yudkowsky.
\newblock The ethics of artificial intelligence.
\newblock \emph{The Cambridge Handbook of Artificial Intelligence}, 2014.

\bibitem[Brockman et~al.(2016)Brockman, Cheung, Pettersson, Schneider,
  Schulman, Tang, and Zaremba]{brockman2016openai}
Greg Brockman, Vicki Cheung, Ludwig Pettersson, Jonas Schneider, John Schulman,
  Jie Tang, and Wojciech Zaremba.
\newblock Openai gym.
\newblock \emph{arXiv preprint arXiv:1606.01540}, 2016.

\bibitem[Buciluǎ et~al.(2006)Buciluǎ, Caruana, and
  Niculescu-Mizil]{buciluǎ2006model}
Cristian Buciluǎ, Rich Caruana, and Alexandru Niculescu-Mizil.
\newblock Model compression.
\newblock In \emph{Proceedings of the 12th ACM SIGKDD international conference
  on Knowledge discovery and data mining}. ACM, 2006.

\bibitem[Carton et~al.(2016)Carton, Helsby, Joseph, Mahmud, Park, Walsh, Cody,
  Patterson, Haynes, and Ghani]{carton2016identifying}
Samuel Carton, Jennifer Helsby, Kenneth Joseph, Ayesha Mahmud, Youngsoo Park,
  Joe Walsh, Crystal Cody, CPT~Estella Patterson, Lauren Haynes, and Rayid
  Ghani.
\newblock Identifying police officers at risk of adverse events.
\newblock In \emph{ACM SIGKDD International Conference on Knowledge Discovery
  and Data Mining}. ACM, 2016.

\bibitem[Chang et~al.(2009)Chang, Boyd-Graber, Gerrish, Wang, and
  Blei]{chang2009reading}
Jonathan Chang, Jordan~L Boyd-Graber, Sean Gerrish, Chong Wang, and David~M
  Blei.
\newblock Reading tea leaves: How humans interpret topic models.
\newblock In \emph{NIPS}, 2009.

\bibitem[Chater and Oaksford(2006)]{chater2006speculations}
Nick Chater and Mike Oaksford.
\newblock Speculations on human causal learning and reasoning.
\newblock \emph{Information sampling and adaptive cognition}, 2006.

\bibitem[Doshi-Velez et~al.(2014)Doshi-Velez, Ge, and
  Kohane]{doshi2014comorbidity}
Finale Doshi-Velez, Yaorong Ge, and Isaac Kohane.
\newblock Comorbidity clusters in autism spectrum disorders: an electronic
  health record time-series analysis.
\newblock \emph{Pediatrics}, 133\penalty0 (1):\penalty0 e54--e63, 2014.

\bibitem[Doshi-Velez et~al.(2015)Doshi-Velez, Wallace, and
  Adams]{doshi2014graph}
Finale Doshi-Velez, Byron Wallace, and Ryan Adams.
\newblock Graph-sparse lda: a topic model with structured sparsity.
\newblock \emph{Association for the Advancement of Artificial Intelligence},
  2015.

\bibitem[Dwork et~al.(2012)Dwork, Hardt, Pitassi, Reingold, and
  Zemel]{Dwork2012}
Cynthia Dwork, Moritz Hardt, Toniann Pitassi, Omer Reingold, and Richard Zemel.
\newblock Fairness through awareness.
\newblock In \emph{Innovations in Theoretical Computer Science Conference}.
  ACM, 2012.

\bibitem[Freitas(2014)]{freitas2014comprehensible}
Alex Freitas.
\newblock Comprehensible classification models: a position paper.
\newblock \emph{ACM SIGKDD Explorations}, 2014.

\bibitem[Garg and Kalai(2016)]{garg2016meta}
Vikas~K Garg and Adam~Tauman Kalai.
\newblock Meta-unsupervised-learning: A supervised approach to unsupervised
  learning.
\newblock \emph{arXiv preprint arXiv:1612.09030}, 2016.

\bibitem[Glennan(2002)]{glennan2002rethinking}
Stuart Glennan.
\newblock Rethinking mechanistic explanation.
\newblock \emph{Philosophy of science}, 2002.

\bibitem[Goodman and Flaxman(2016)]{goodman2016european}
Bryce Goodman and Seth Flaxman.
\newblock European union regulations on algorithmic decision-making and a"
  right to explanation".
\newblock \emph{arXiv preprint arXiv:1606.08813}, 2016.

\bibitem[Gupta et~al.(2016)Gupta, Cotter, Pfeifer, Voevodski, Canini, Mangylov,
  Moczydlowski, and Van~Esbroeck]{gupta2016monotonic}
Maya Gupta, Andrew Cotter, Jan Pfeifer, Konstantin Voevodski, Kevin Canini,
  Alexander Mangylov, Wojciech Moczydlowski, and Alexander Van~Esbroeck.
\newblock Monotonic calibrated interpolated look-up tables.
\newblock \emph{Journal of Machine Learning Research}, 2016.

\bibitem[Hamill(2017)]{cmupoker}
Sean Hamill.
\newblock {CMU} computer won poker battle over humans by statistically
  significant margin.
\newblock \url{
  http://www.post-gazette.com/business/tech-news/2017/01/31/CMU-computer-won-poker-battle-over-humans-by-statistically-significant-margin/stories/201701310250
  }, 2017.
\newblock Accessed: 2017-02-07.

\bibitem[Hardt and Talwar(2010)]{Hardt2010}
Moritz Hardt and Kunal Talwar.
\newblock On the geometry of differential privacy.
\newblock In \emph{ACM Symposium on Theory of Computing}. ACM, 2010.

\bibitem[Hardt et~al.(2016)Hardt, Price, and Srebro]{hardt2016equality}
Moritz Hardt, Eric Price, and Nati Srebro.
\newblock Equality of opportunity in supervised learning.
\newblock In \emph{Advances in Neural Information Processing Systems}, 2016.

\bibitem[Hempel and Oppenheim(1948)]{hempel1948studies}
Carl Hempel and Paul Oppenheim.
\newblock Studies in the logic of explanation.
\newblock \emph{Philosophy of science}, 1948.

\bibitem[Ho and Basu(2002)]{ho2002complexity}
Tin~Kam Ho and Mitra Basu.
\newblock Complexity measures of supervised classification problems.
\newblock \emph{IEEE transactions on pattern analysis and machine
  intelligence}, 2002.

\bibitem[Keil(2006)]{keil2006explanation}
Frank Keil.
\newblock Explanation and understanding.
\newblock \emph{Annu. Rev. Psychol.}, 2006.

\bibitem[Keil et~al.(2004)Keil, Rozenblit, and Mills]{keil2004lies}
Frank Keil, Leonid Rozenblit, and Candice Mills.
\newblock What lies beneath? understanding the limits of understanding.
\newblock \emph{Thinking and seeing: Visual metacognition in adults and
  children}, 2004.

\bibitem[Kim et~al.(2013)Kim, Chacha, and Shah]{kim2013inferring}
Been Kim, Caleb Chacha, and Julie Shah.
\newblock Inferring robot task plans from human team meetings: A generative
  modeling approach with logic-based prior.
\newblock \emph{Association for the Advancement of Artificial Intelligence},
  2013.

\bibitem[Kim et~al.(2015{\natexlab{a}})Kim, Glassman, Johnson, and
  Shah]{kim2015ibcm}
Been Kim, Elena Glassman, Brittney Johnson, and Julie Shah.
\newblock i{BCM}: Interactive bayesian case model empowering humans via
  intuitive interaction.
\newblock 2015{\natexlab{a}}.

\bibitem[Kim et~al.(2015{\natexlab{b}})Kim, Shah, and Doshi-Velez]{kim2015mind}
Been Kim, Julie Shah, and Finale Doshi-Velez.
\newblock Mind the gap: A generative approach to interpretable feature
  selection and extraction.
\newblock In \emph{Advances in Neural Information Processing Systems},
  2015{\natexlab{b}}.

\bibitem[Lakkaraju et~al.(2016)Lakkaraju, Bach, and
  Leskovec]{lakkaraju2016interpretable}
Himabindu Lakkaraju, Stephen~H Bach, and Jure Leskovec.
\newblock Interpretable decision sets: A joint framework for description and
  prediction.
\newblock In \emph{Proceedings of the 22nd ACM SIGKDD International Conference
  on Knowledge Discovery and Data Mining}, pages 1675--1684. ACM, 2016.

\bibitem[Lazar et~al.(2010)Lazar, Feng, and Hochheiser]{lazar2010research}
Jonathan Lazar, Jinjuan~Heidi Feng, and Harry Hochheiser.
\newblock \emph{Research methods in human-computer interaction}.
\newblock John Wiley \& Sons, 2010.

\bibitem[Lei et~al.(2016)Lei, Barzilay, and Jaakkola]{lei2016rationalizing}
Tao Lei, Regina Barzilay, and Tommi Jaakkola.
\newblock Rationalizing neural predictions.
\newblock \emph{arXiv preprint arXiv:1606.04155}, 2016.

\bibitem[Lombrozo(2006)]{lombrozo2006structure}
Tania Lombrozo.
\newblock The structure and function of explanations.
\newblock \emph{Trends in cognitive sciences}, 10\penalty0 (10):\penalty0
  464--470, 2006.

\bibitem[Lou et~al.(2012)Lou, Caruana, and Gehrke]{lou2012intelligible}
Yin Lou, Rich Caruana, and Johannes Gehrke.
\newblock Intelligible models for classification and regression.
\newblock In \emph{ACM SIGKDD international conference on Knowledge discovery
  and data mining}. ACM, 2012.

\bibitem[Mnih et~al.(2013)Mnih, Kavukcuoglu, Silver, Graves, Antonoglou,
  Wierstra, and Riedmiller]{mnih2013playing}
Volodymyr Mnih, Koray Kavukcuoglu, David Silver, Alex Graves, Ioannis
  Antonoglou, Daan Wierstra, and Martin Riedmiller.
\newblock Playing atari with deep reinforcement learning.
\newblock \emph{arXiv preprint arXiv:1312.5602}, 2013.

\bibitem[Neath and Surprenant(2003)]{neath2003human}
Ian Neath and Aimee Surprenant.
\newblock \emph{Human Memory}.
\newblock 2003.

\bibitem[Otte(2013)]{otte2013safe}
Clemens Otte.
\newblock Safe and interpretable machine learning: A methodological review.
\newblock In \emph{Computational Intelligence in Intelligent Data Analysis}.
  Springer, 2013.

\bibitem[Parliament and of~the European~Union(2016)]{eu_reg}
Parliament and Council of~the European~Union.
\newblock General data protection regulation.
\newblock 2016.

\bibitem[Ribeiro et~al.(2016)Ribeiro, Singh, and Guestrin]{ribeiro2016should}
Marco~Tulio Ribeiro, Sameer Singh, and Carlos Guestrin.
\newblock ``why should i trust you?": Explaining the predictions of any
  classifier.
\newblock \emph{arXiv preprint arXiv:1602.04938}, 2016.

\bibitem[Ruggieri et~al.(2010)Ruggieri, Pedreschi, and
  Turini]{ruggieri2010data}
Salvatore Ruggieri, Dino Pedreschi, and Franco Turini.
\newblock Data mining for discrimination discovery.
\newblock \emph{ACM Transactions on Knowledge Discovery from Data}, 2010.

\bibitem[Schulz et~al.(2016)Schulz, Tenenbaum, Duvenaud, Speekenbrink, and
  Gershman]{schulz2016compositional}
Eric Schulz, Joshua Tenenbaum, David Duvenaud, Maarten Speekenbrink, and Samuel
  Gershman.
\newblock Compositional inductive biases in function learning.
\newblock \emph{bioRxiv}, 2016.

\bibitem[Sculley et~al.(2015)Sculley, Holt, Golovin, Davydov, Phillips, Ebner,
  Chaudhary, Young, Crespo, and Dennison]{sculley2015hidden}
D~Sculley, Gary Holt, Daniel Golovin, Eugene Davydov, Todd Phillips, Dietmar
  Ebner, Vinay Chaudhary, Michael Young, Jean-Fran{\c{c}}ois Crespo, and Dan
  Dennison.
\newblock Hidden technical debt in machine learning systems.
\newblock In \emph{Advances in Neural Information Processing Systems}, 2015.

\bibitem[Silver et~al.(2016)Silver, Huang, Maddison, Guez, Sifre, Van
  Den~Driessche, Schrittwieser, Antonoglou, Panneershelvam, Lanctot,
  et~al.]{silver2016mastering}
David Silver, Aja Huang, Chris~J Maddison, Arthur Guez, Laurent Sifre, George
  Van Den~Driessche, Julian Schrittwieser, Ioannis Antonoglou, Veda
  Panneershelvam, Marc Lanctot, et~al.
\newblock Mastering the game of go with deep neural networks and tree search.
\newblock \emph{Nature}, 2016.

\bibitem[Strahilevitz(2008)]{strahilevitz2008privacy}
Lior~Jacob Strahilevitz.
\newblock Privacy versus antidiscrimination.
\newblock \emph{University of Chicago Law School Working Paper}, 2008.

\bibitem[Suissa-Peleg et~al.(2016)Suissa-Peleg, Haehn, Knowles-Barley, Kaynig,
  Jones, Wilson, Schalek, Lichtman, and Pfister]{suissa2016automatic}
Adi Suissa-Peleg, Daniel Haehn, Seymour Knowles-Barley, Verena Kaynig, Thouis~R
  Jones, Alyssa Wilson, Richard Schalek, Jeffery~W Lichtman, and Hanspeter
  Pfister.
\newblock Automatic neural reconstruction from petavoxel of electron microscopy
  data.
\newblock \emph{Microscopy and Microanalysis}, 2016.

\bibitem[Toubiana et~al.(2010)Toubiana, Narayanan, Boneh, Nissenbaum, and
  Barocas]{toubiana2010adnostic}
Vincent Toubiana, Arvind Narayanan, Dan Boneh, Helen Nissenbaum, and Solon
  Barocas.
\newblock Adnostic: Privacy preserving targeted advertising.
\newblock 2010.

\bibitem[Vanschoren et~al.(2014)Vanschoren, Van~Rijn, Bischl, and
  Torgo]{vanschoren2014openml}
Joaquin Vanschoren, Jan~N Van~Rijn, Bernd Bischl, and Luis Torgo.
\newblock Openml: networked science in machine learning.
\newblock \emph{ACM SIGKDD Explorations Newsletter}, 15\penalty0 (2):\penalty0
  49--60, 2014.

\bibitem[Varshney and Alemzadeh(2016)]{VarshneyA16}
Kush Varshney and Homa Alemzadeh.
\newblock On the safety of machine learning: Cyber-physical systems, decision
  sciences, and data products.
\newblock \emph{CoRR}, 2016.

\bibitem[Wang and Rudin(2015)]{wang2015falling}
Fulton Wang and Cynthia Rudin.
\newblock Falling rule lists.
\newblock In \emph{AISTATS}, 2015.

\bibitem[Wang et~al.(2017)Wang, Rudin, Doshi-Velez, Liu, Klampfl, and
  MacNeille]{bayesian2017wang}
Tong Wang, Cynthia Rudin, Finale Doshi-Velez, Yimin Liu, Erica Klampfl, and
  Perry MacNeille.
\newblock Bayesian rule sets for interpretable classification.
\newblock In \emph{International Conference on Data Mining}, 2017.

\bibitem[Williams et~al.(2016)Williams, Kim, Rafferty, Maldonado, Gajos,
  Lasecki, and Heffernan]{williams2016axis}
Joseph~Jay Williams, Juho Kim, Anna Rafferty, Samuel Maldonado, Krzysztof~Z
  Gajos, Walter~S Lasecki, and Neil Heffernan.
\newblock Axis: Generating explanations at scale with learnersourcing and
  machine learning.
\newblock In \emph{ACM Conference on Learning@ Scale}. ACM, 2016.

\bibitem[Wilson et~al.(2015)Wilson, Dann, Lucas, and Xing]{wilson2015human}
Andrew Wilson, Christoph Dann, Chris Lucas, and Eric Xing.
\newblock The human kernel.
\newblock In \emph{Advances in Neural Information Processing Systems}, 2015.

\end{thebibliography}

\end{document}